\def\year{2019}\relax
\documentclass[letterpaper]{article} 
\usepackage{aaai19}  
\usepackage{times}  
\usepackage{helvet}  
\usepackage{courier}  
\usepackage{url}  
\usepackage{graphicx}  

\usepackage{booktabs} 
\usepackage{xcolor}
\usepackage{subcaption}
\usepackage{xspace}

\usepackage{amsmath}
\usepackage{bbold}
\usepackage{amsfonts}
\usepackage{amsthm}
\newtheorem{definition}{Definition}

\begin{document}

\title{Putting Fairness Principles into Practice: Challenges, Metrics, and Improvements}

\author{\\[-5mm]\textbf{Alex Beutel, Jilin Chen, Tulsee Doshi, Hai Qian, Allison Woodruff, Christine Luu}, \\
\textbf{Pierre Kreitmann, Jonathan Bischof, Ed H. Chi}\\
\{alexbeutel, jilinc, tulsee, hqian, woodruff, cmluu, kreitmann, bischof, edchi\}@google.com\\
Google
}
\nocopyright
\maketitle

\newcommand{\conditional}{conditional equality\xspace}
\newcommand{\fprgap}{FPR gap\xspace}
\newcommand{\user}{user\xspace}
\newcommand{\users}{users\xspace}
\newcommand{\mindiff}{absolute correlation\xspace}
\newcommand{\mindifflossshort}{corr. reg.\xspace}
\newcommand{\mindiffloss}{absolute correlation regularization\xspace}
\newcommand{\mlfairness}{algorithmic fairness\xspace}
\newcommand{\groupone}{Group 1\xspace}
\newcommand{\grouptwo}{Group 2\xspace}
\newcommand{\notgroupone}{Not-\groupone}
\newcommand{\notgrouptwo}{Not-\grouptwo}
\newcommand{\ratio}[1]{$#1\times$\xspace}
\newcommand{\myparagraph}[1]{\vspace{1mm}\noindent \textbf{#1} }
\newcommand{\negset}{{\mathcal{X}^-}}

\newcommand{\citet}[1]{\citeauthor{#1}~\shortcite{#1}}

\begin{abstract}
As more researchers have become aware of and passionate about \mlfairness, there has been an explosion in papers laying out new metrics, suggesting algorithms to address issues, and calling attention to issues in existing applications of machine learning.  This research has greatly expanded our understanding of the concerns and challenges in deploying machine learning, but there has been much less work in seeing how the rubber meets the road.

In this paper we provide a case-study on the application of fairness in machine learning research to a production classification system, and offer new insights in how to measure and address \mlfairness issues.  We discuss open questions in implementing equality of opportunity and describe our fairness metric, \emph{\conditional}, that takes into account distributional differences.  Further, we provide a new approach to improve on the fairness metric during model training and demonstrate its efficacy in improving performance for a real-world product.
\end{abstract}

\maketitle

\section{Introduction}

By almost every measure, there has been an explosion in attention and research on machine learning fairness: there is a quickly growing amount of research on how to define, measure, and address machine learning fairness, and products are evaluated with these concerns in mind.
Despite this significant attention, there has been much less published work detailing how fairness concerns are measured and addressed by product teams in industry.  In this paper, we hope to shed light on the challenges in following these principles and learnings in an applied production setting, and to offer metrics and methods developed in the process.

We focus on a classification system where adverse actions are taken against examples predicted to be in the positive class.  This is similar to not giving a person a mortgage if a model predicts they will default on it \cite{hardt2016equality}, using recidivism prediction for setting bail \cite{chouldechova2017fair}, or removing comments on the web if they are predicted to be abusive \cite{dixon2017measuring}.  In all of these cases, each item is associated with a user, and if the classifier makes a mistake and the adverse action is taken against their example, that is bad for the user. 
More generally, if examples from certain groups of users more often have adverse actions taken against them, it could effect the health of the service.  As a result, improving group fairness \cite{hardt2016equality} is both the right thing to do and important to the health of the product.

\begin{figure}[t]
    \centering
    \begin{subfigure}[b]{0.45\columnwidth}
        \includegraphics[width=\textwidth]{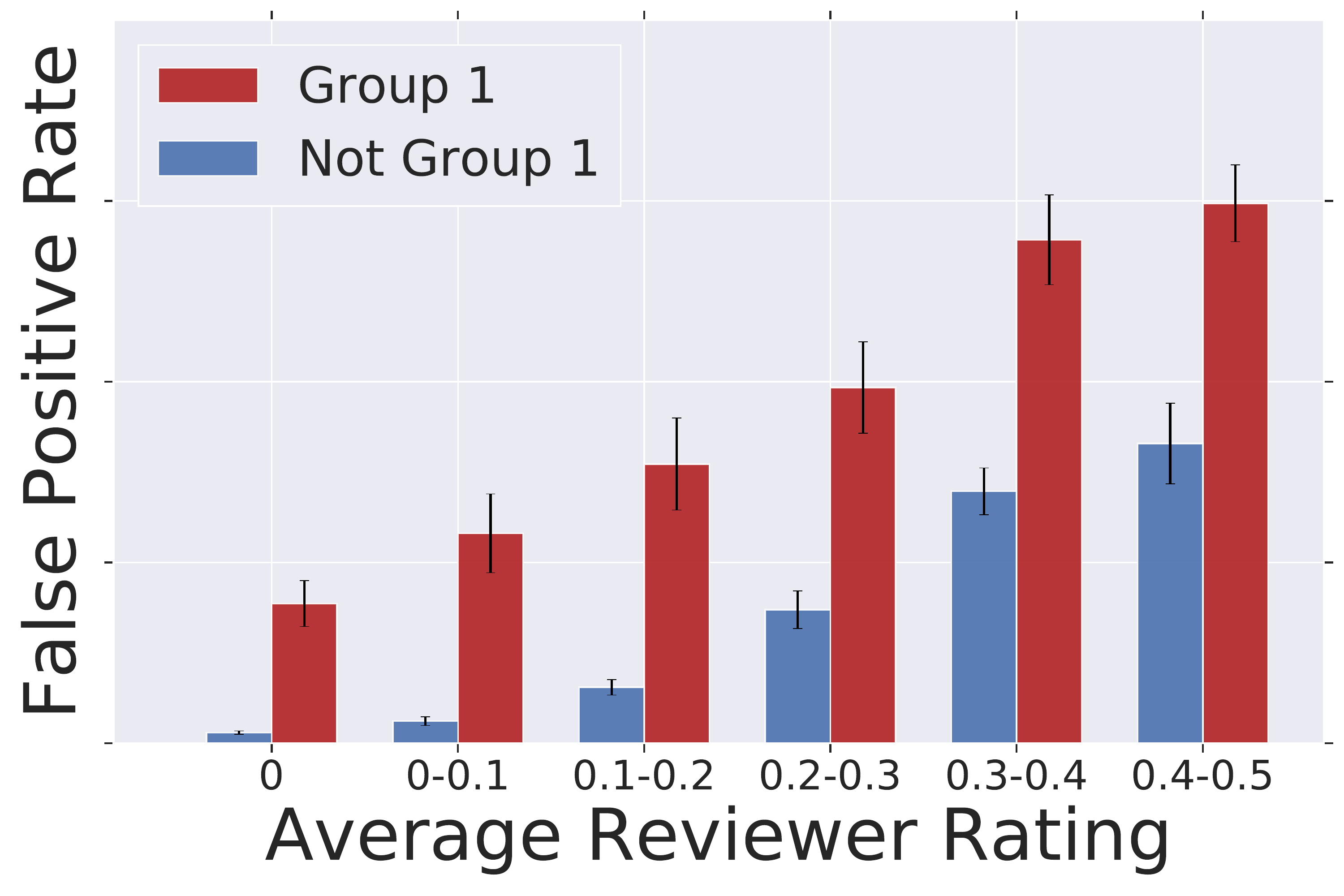}
        \caption{Before}
    \end{subfigure}
    \begin{subfigure}[b]{0.45\columnwidth}
        \includegraphics[width=\textwidth]{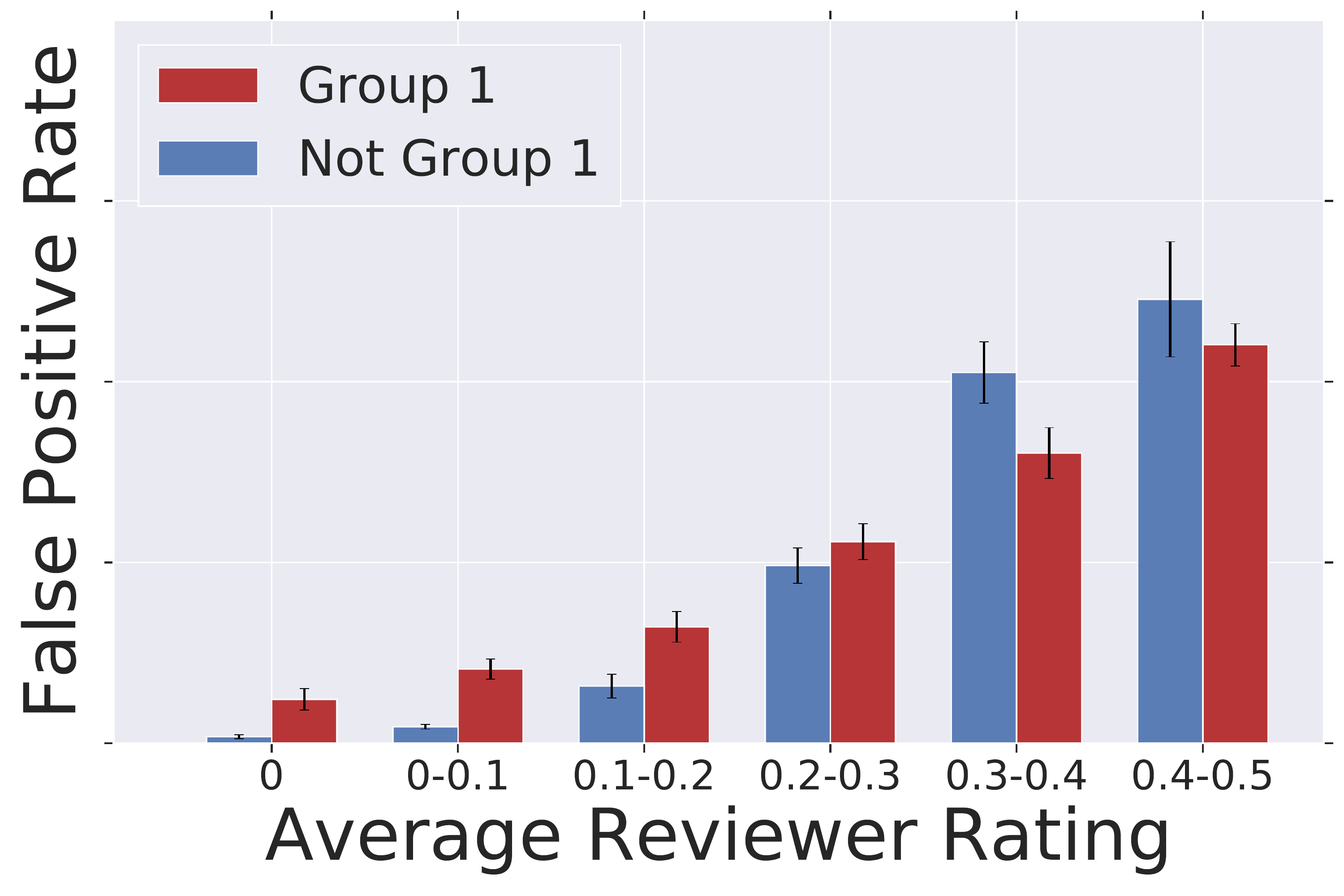}
        \caption{After Improvements}
    \end{subfigure}
    \caption{We observe a significant improvement in the gap in false positive rate by training the model with \mindiffloss.  
    }
    \label{fig:crownjewel}
\end{figure}

We focus on equality of opportunity \cite{hardt2016equality}, in particular comparing false positive rate (FPR) between groups.  While the model being calibrated \cite{crowson2016assessing} is an important mathematical property, it does not reflect the experience of \users and the implications of representation on the service.  However, while \cite{hardt2016equality} provides great intuition and philosophical guidance, we find that in practice it leaves significant wiggle room in how the metric is calculated based on how the evaluation data is sampled or generated.  
Further, as shown by \citet{corbett2017algorithmic}, distributional differences can result in unintended side-effects and costs when implementing fairness changes.   We address these issues through a generalized form of the metric, \emph{\conditional}, that makes these decisions more explicit, and we describe how we navigated these challenges in our use case. Figure \ref{fig:crownjewel} presents a summary of our results in an applied production setting\footnote{Due to the sensitive nature of these tests, we must omit the numerical values on the y-axis of all plots.  In all cases, plots that are juxtaposed keep the same range on the y-axis such that results can be compared.}.

Given this metric, we consider how to improve it under the practical constraints of a product.  For example, we are unable to reliably observe the sensitive attribute at inference time, preventing approaches like using different thresholds \cite{lipton2017does}.  Further, as with many engineering systems, simplicity and maintainability are core requirements.  We begin with exploring the use of adversarial training techniques \cite{EOadv,DBLP:journals/corr/abs-1801-07593,laftr}, which have been shown to be effective.  However, as with many adversarial training approaches, we find these are sometimes unstable and difficult to reliably train well.  As a result, we offer a new approach, \emph{\mindiffloss}, which, while not provably optimal at convergence, we find empirically can stably improve our algorithmic fairness metrics.  

Finally, we test these approaches on the production model to improve the metrics for 
items from two sensitive groups.
We  find that adversarial training and \mindiffloss both improve these metrics significantly.

The open questions of how to design a practical \mlfairness metric and how to improve that metric under the constraints of a production system are crucial to putting academic learning into practice in industry.  While our approaches are tailored to the application and constraints at hand, we believe they can offer guidance to ML practitioners and call attention to gaps in the current literature that researchers can work to address and practitioners should be mindful of.
We list below our contributions:
\begin{enumerate}
    \item \textbf{Metrics:} We demonstrate the challenge in ``correctly'' measuring equality of opportunity, and describe our metric, \conditional, that makes practitioner decisions explicit  and takes into account varying difficulty of examples across groups. 
    \item \textbf{Optimization:} We offer a new regularization technique, called \mindiffloss, to encourage equality of opportunity during training.
    \item \textbf{Improvement:} We demonstrate improvements to our \mlfairness metrics.  In particular, we find that traditional modelling, such as through larger models, can improve \mlfairness.  Second, we find that \mindiffloss stably and significantly improves \mlfairness metrics.
\end{enumerate}

\section{Background and Related Work}
We begin with some background material on \mlfairness metrics and relevant related work.

\myparagraph{Metrics}
Many different metrics have been proposed to measure machine learning fairness, particularly for binary classification.  One line of work called individual fairness rests on the view that similar examples should receive similar predictions \cite{fairaware}; but this leaves open the question of similarity.
Another line of work focuses on group fairness, where examples are grouped by a particular sensitive attribute and statistics about the model predictions are aggregated within the group and compared between groups.  Multiple group fairness metrics have been proposed.  Demographic parity \cite{calders2010three} asserts that the average prediction for each group should be equal:
\begin{align}
P(\hat{y}=1 | s=0) = P(\hat{y} = 1 | s=1),
\end{align}
where the model prediction is $\hat{y}$ and the sensitive attribute (group identity) is given by $s$.  One issue with this view is that different groups could have very different labels $y$ (often called different base rates).  Equality of opportunity \cite{hardt2016equality} addresses this by analyzing the accuracy and asserting that the model should not mistake $y=0$ examples for $y=1$ examples at a higher rate for one group than another:
\begin{align}
    P(\hat{y}=0| s=0, y=0) = P(\hat{y}=0 | s=1, y=0)
\end{align}
Empirically this means that we are comparing the false positive rate (FPR) for examples from each group, which makes sense if a false positives result in a high cost to the group.  A symmetric statement can be made for the false negative rate, and putting these together is defined as equality of odds.
A third popular group fairness metric has been calibration \cite{crowson2016assessing}:
\begin{align}
    E[y | s = 0, \hat{y} = p] = E[y | s = 1, \hat{y} = p] \;\; \forall p \in [0,1]
\end{align}
Significant work has analyzed these metrics and their gaps.  A number of results have shown that achieving all of them (or even pairs of them) is only possible in limited cases \cite{kleinberg2016inherent,DBLP:conf/nips/PleissRWKW17}.
Other research has considered expanding them to more complex combinations of multiple sensitive attributes, which we refer to as intersectional testing \cite{DBLP:journals/corr/abs-1711-05144,DBLP:journals/corr/abs-1711-08513}. 
Another different line of work has explored using the language of causality to define fairness \cite{DBLP:conf/nips/KilbertusRPHJS17}, but this has had limited traction \cite{garg2018counterfactual} due to the difficulty of knowing the causal graph.

\myparagraph{Modeling}
With this wide variety of measures of fairness, another line of research has explored how to address \mlfairness issues in models.  One line of work has built on adversarial training.  This approach began for domain adaptation \cite{DBLP:journals/corr/AjakanGLLM14,ganin2016domain,bousmalis2016domain} and was quickly applied to fairness \cite{zemel2013learning,louizos2015variational,edwards2015censoring}.  More recent work has modified this to align with different ML fairness metrics \cite{EOadv,DBLP:journals/corr/abs-1801-07593,laftr}.
Others have focused on constrained optimization \cite{DBLP:journals/corr/abs-1803-02453,DBLP:conf/nips/GohCGF16}, or using a variety of regularization techniques \cite{kamishima2011fairness,zafar2015fairness,bechavod2017penalizing}.  Our regularization approach draws  from all of these bodies of work.

Other approaches have been advocated for such as using different thresholds for each group during binary classification \cite{lipton2017does}, but this is not feasible without observing the sensitive attribute at inference.    
Another approach has been data augmentation \cite{dixon2017measuring}, but it is often unclear how to do this in applications with more complex, arbitrary feature sets, e.g., loan default prediction over individuals or recommender systems where domain adaptation is difficult.

\myparagraph{Application:} 
Much of the published work on addressing fairness concerns focuses on public policy applications like recidivism prediction \cite{chouldechova2017fair,lum2017limitations}, predictive policing \cite{lum2016predict}, and child services \cite{chouldechova2018case}.  These works explore some related practical difficulties, e.g., \cite{chouldechova2017fair} discusses how metrics could be calculated after conditioning on other covariates like prior convictions.  Recently, \citet{holstein2018improving} surveyed practitioners on the challenges to improving fairness in industry.

\section{Application Setting}
\label{sec:app_setting}
We begin with an overview of our application, and the properties of it that are key to how we define and address any fairness concerns.  We focus on a binary classification model that predicts if each example follows or breaks  a pre-determined product policy. Examples that break the policy have an adverse action taken directly against them; examples that fall within policy have no action taken against them.  This is similar to abuse classification literature \cite{dixon2017measuring}, the common loan-default prediction problem \cite{hardt2016equality}, or recidivism prediction \cite{chouldechova2017fair}.

Because of the quantity of data, reviewers cannot rate all examples on the service.  Rather, we use human raters to score a subsample of the examples.  Human raters give a score $y \in [0,1]$ and $K$ raters score each example, producing an average ground truth score $y_i = \sum_k \frac{y_{i,k}}{K}$ for example $i$.  We can choose which examples to get rated, but each rating is relatively expensive such that we can only have a small fraction of the data rated.  This is particularly exacerbated by the fact that only a small percentage of the examples break policy and as such, random sampling of examples produces relatively little data with high $y$.

Because most of the examples are unrated, we use a model $f$ to predict the ground truth score: $f(\mathbf{x}_i) = \hat{y}_i \approx y_i$ where $\mathbf{x} \in \mathbb{R}^d$ are features of the example.  We take $\mathbf{x}$ to be a general feature vector; in practice, it contains a wide range of features including direct features as well as embedding and signals produced from other models and systems. 
The model is trained as a regression model with squared error: $\min_f \sum_i (f(\mathbf{x}_i) - y_i)^2$.  At inference time, adverse action is taken against all examples with prediction over some threshold: $f(\mathbf{x}_i) = \hat{y}_i > \tau$.

For evaluating \mlfairness, we consider group fairness \cite{hardt2016equality} over groups of examples or \users.  Each example is associated with a \user, but features about the \users are not reliably observed, i.e., we consider the case where \users are not associated with demographic information.   Rather, a small number of \users are willing to share their demographics, which we can use during training and for offline evaluation of \mlfairness metrics.  We refer to the group membership by feature $s$.  Because most \users do not share their demographic information, we cannot use those features as input to the model or in any way at inference time.  Further, because the number of examples with demographic information is relatively small, a more expansive intersectional evaluation is not feasible \cite{DBLP:journals/corr/abs-1711-05144,DBLP:journals/corr/abs-1711-08513}.

\paragraph{Assumptions} We take the product policy as ground truth. Further, we make the simplifying assumption that the human raters provide an unbiased estimate of the ground truth score.  At the present, this is difficult to evaluate and further research is needed on how to detect and evaluate rater bias.  We offer an expanded discussion of the assumptions and limitations of our analysis
at the end of the paper.

\subsection{Baseline Model}
We consider now how our baseline model performs, in particular the FPR.  This original model is a linear model over diverse feature set $\mathbf{x}$.
We consider the FPR for two important sensitive groups, which we will refer to as \groupone and \grouptwo.  In each case, we compare the FPR to that of examples not from that respective \user group, i.e., \notgroupone and \notgrouptwo.
Due to the sensitive nature of the measurements, we present all results in relativistic terms.  For example, the FPR Ratio is defined as:
\begin{align}
\mbox{FPR Ratio}_{\rm \groupone} = \frac{{\rm FPR}_{\rm \groupone}}{{\rm FPR}_{\rm \notgroupone}}
\end{align}

We present these measurements of the original system in Figure \ref{fig:linear_groupone} and \ref{fig:linear_grouptwo}.  As observed in the plots, we find that for both \groupone and \grouptwo the FPR Ratio $> 1$, with FPR Ratio$_{\rm \groupone} = 5.3\times$ and FPR Ratio$_{\rm \grouptwo} = 2.18\times$. 
Both of these numbers indicate a gap in the equality of opportunity and that examples from these groups are more frequently incorrectly having adverse action taken against them.  
Although hard to measure due to limited data, we generally observe lower FNR for subgroups than the majority, and because adverse actions are taken for predicted positives, we focus on the FPR in the rest of our analysis.

\section{ML Fairness Metric}
While equality of opportunity \cite{hardt2016equality} provides insight into measuring cost to \users of mistakes across groups, it leaves many open questions that in practice need to be addressed.  In particular, how should the data that the metric is calculated over be sampled?  What if the distribution of data differs?  How do we address these differences?

\subsection{Data Distribution}
One immediate open question in the analysis and practical implementation is: \emph{how should the data be sampled?}  FPR and FNR are \emph{only} meaningful with respect to a given distribution of evaluation data, and we find that different ways of generating that evaluation data give significantly different results.  Further, as pointed out by \citet{corbett2017algorithmic}, examples from different groups may have different distributions of risk, and directly applying equality of opportunity over these different risks can raise its own issues.

\begin{figure*}
    \centering
    \begin{subfigure}[b]{0.28\textwidth}
        \includegraphics[width=\textwidth]{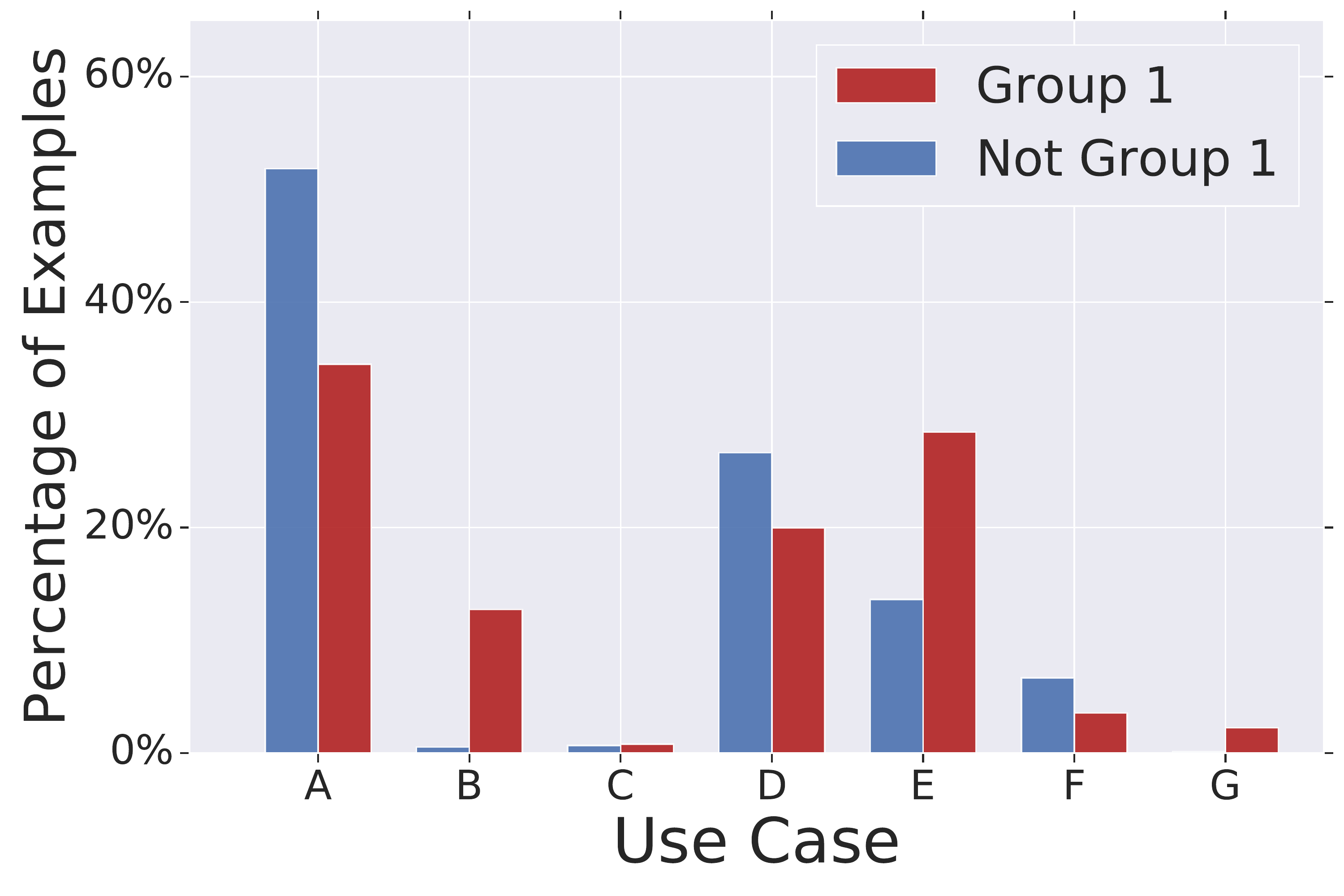}
        \caption{Distribution of Use Cases}
        \label{fig:pdf_usecases}
    \end{subfigure}
    \begin{subfigure}[b]{0.28\textwidth}
        \includegraphics[width=\textwidth]{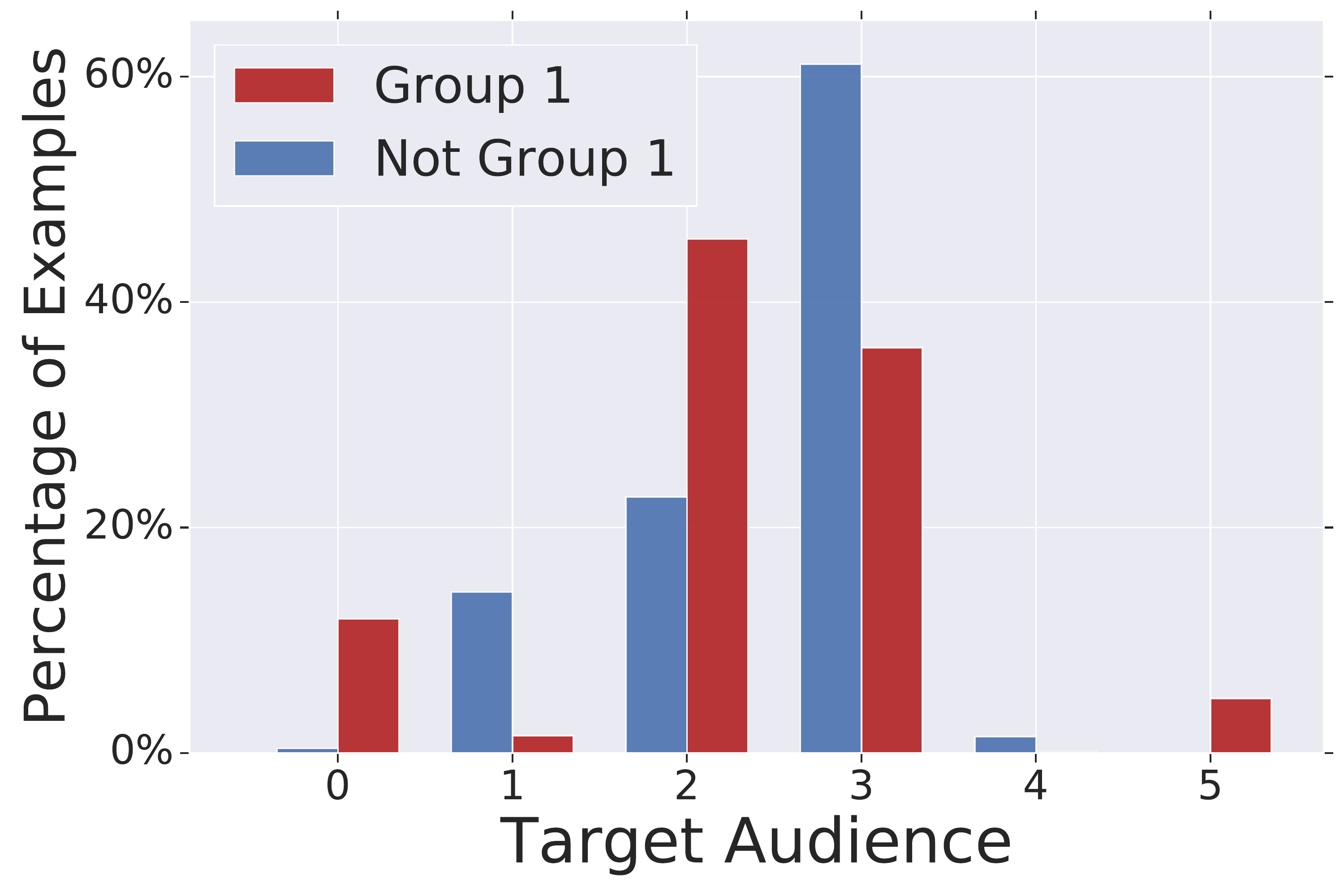}
        \caption{Distribution of Audience}
        \label{fig:pdf_audience}
    \end{subfigure}
    \begin{subfigure}[b]{0.28\textwidth}
        \includegraphics[width=\textwidth]{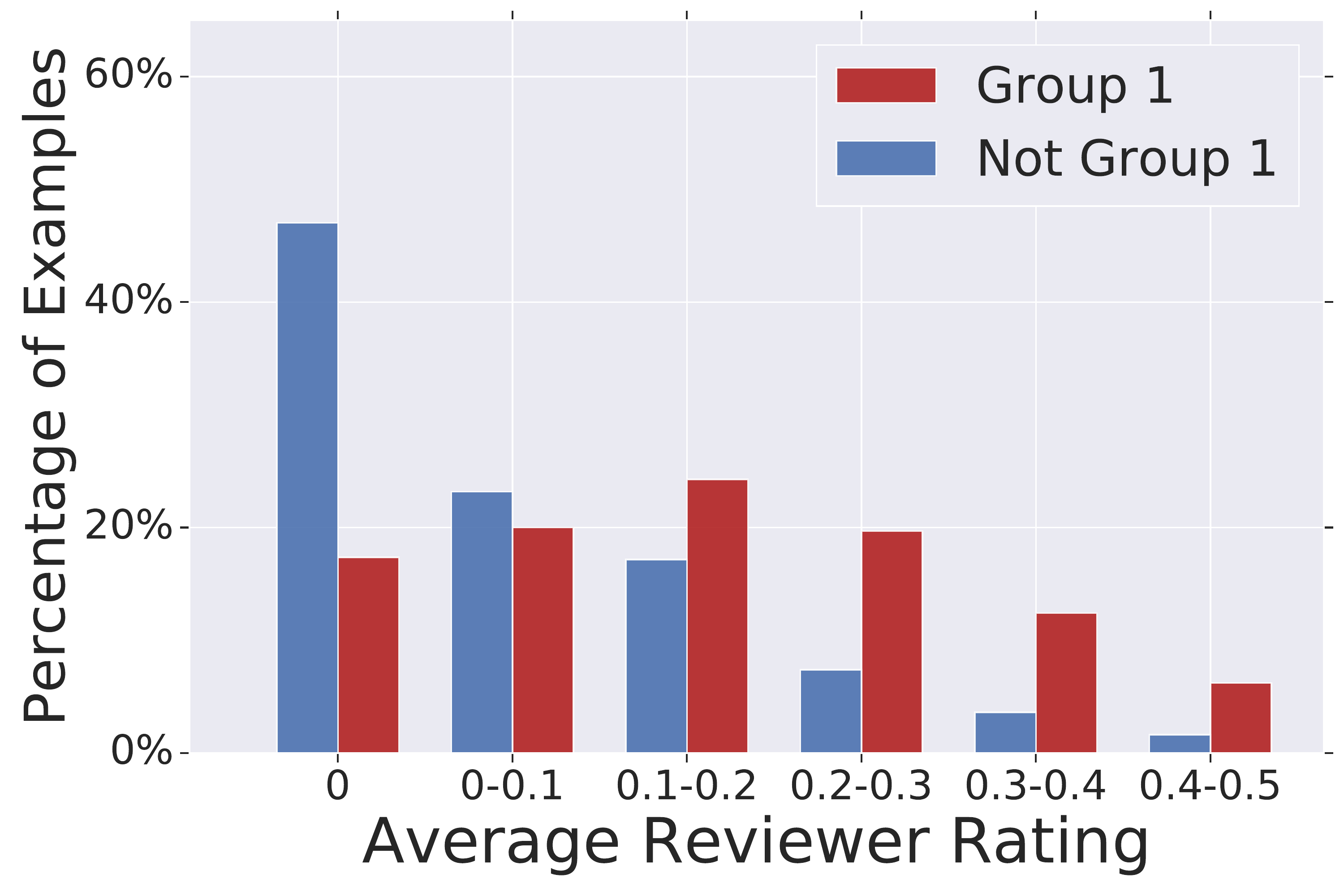}
        \caption{Distribution of Ratings $y$}
        \label{fig:pdf_ratings}
    \end{subfigure}
    \caption{Different groups have different distribution of examples.}
    \label{fig:pdfs}
\end{figure*}

The analysis above is based on a dataset built by sampling examples proportional to their usage,
but ignore many other differences in the data.  Unsurprisingly, different groups of \users are associated with different types of examples, such as with different use cases or target audiences.
For example, in Figures \ref{fig:pdf_usecases} and \ref{fig:pdf_audience} we find that for one particular demographic property, the distribution of both use cases of the examples as well as the target audiences are quite different between the groups (considered over negative examples from each group).

While the data can be stratified by many of these dimensions, no principled way has been given for how or when to do so. 
Here, we take inspiration from \citet{corbett2017algorithmic}, which suggests the importance of addressing different risk distributions.  However, \citet{corbett2017algorithmic} analyze risk through the model's predictions rather than through some externally observable property.  Here, we deviate in that we observe a real-valued policy $y$ averaged over multiple human raters.  As can be seen in Figure \ref{fig:pdf_ratings}, we find that even within examples that would be considered negatives ($y < \tau$), there is a notably different distribution between groups.  In particular, the sensitive subgroup has relatively more examples close to the policy threshold $\tau$, suggesting uncertainty by human raters about how the examples align with the policy.

\subsection{Distribution-Dependent Metrics}
Understanding and addressing these differences in distribution is critical to interpreting the results.  Therefore, we begin with laying out a formalization for conditional group fairness metrics and then discuss the reasoning and implications of our choice of what to condition on.

First we build on \cite{ritov2017conditional} and define a conditional group fairness  for our case:
\begin{definition}
Conditional Equality of Opportunity is defined for conditioning on feature $A$ that takes values $\mathcal{A}$:
\begin{align}
P(\hat{y} \geq \tau &| s=1, y < \tau, A = a)\\&= P(\hat{y} \geq \tau | s=0, y < \tau, A = a) \; \forall a\in\mathcal{A}\notag
\end{align}
\end{definition}
This definition does not give a concrete metric and leaves the question of how to prioritize different $a \in \mathcal{A}$.  We can make this precise by defining the conditional equality of opportunity gap:
\begin{definition}
Conditional equality of opportunity gap is defined conditioning over feature $A$ taking values in $\mathcal{A}$, with each gap weighted by a probability $p_a$ for $a \in \mathcal{A}$:
\begin{align}
    \label{eq:EO_diff}
    EOGap = \sum_a& p_a [P(\hat{y} \geq \tau | s=1, y < \tau , A = a) \\&- P(\hat{y} \geq \tau | s=0, y < \tau, A=a)]\notag
\end{align}
\end{definition}
By setting $p_a = \frac{1}{|\mathcal{A}|}$, this metric equally weights each possible value $a \in \mathcal{A}$.  This is generally a good prior absent a strong reason to deviate.
However, if one group has a skew in $A$ then the uniform prior may not represent the experience for a \user in that group.  In this case, another option is to set $p_a = P(A = a| s = 0)$, which aligns with importance weighting the data from the background distribution to match the distribution of the focused subgroup\footnote{When reporting ratios, we compute the ratio of average FPRs to align with this view of data sampling.}.

Crucially, with this definition we still must choose a feature $A$ on which to condition.  Conditioning on a particular property, for example the example use case, would have the implication that error rates can be different across use cases, as long as they are the same across groups for the same use case; but if different groups prefer different use cases, then this metric would not necessarily support those group preferences. 
As discussed above, we believe the averaged human rating addresses a balance of desired properties for our metric. 
The averaged human rating does not prioritize different use cases, target audiences, etc., but rather we can interpret it as giving us a way of observing the inherent difficulty (or risk as in \cite{corbett2017algorithmic}) of an example.  If humans are uncertain if an example meets a policy, it is understandable for the model to make a mistake as well.  As we see in Figure \ref{fig:crownjewel}, we observe that the FPR does increase with the averaged human rater score, and while we still observe a gap in FPR between the background data and the subgroup, it is partially explained by the difference in the distribution of examples.

Note, these proposals have significant connections to related work. 
Most related is \cite{ritov2017conditional}, which first proposed this generalized fairness metric and included conditioning on arbitrary variables.
The similarity function in individual fairness can be thought of as conditioning on different features, and that work grapples with many of these same issues \cite{dwork2012fairness}.
\citet{chouldechova2017fair} and \citet{corbett2017algorithmic} both mention that metrics can be calculated condition on other variables, and for recidivism prediction they condition on the number of prior convictions, but neither give a general framework for when and how to condition.  In examining fairness through a causal lens, \cite{DBLP:conf/nips/KilbertusRPHJS17} explores the question of what is a ``resolving variable,'' but ultimately this is left as a philosophical choice.  Further, our \conditional metric can be viewed as a special case of the intersectional fairness analysis \cite{DBLP:journals/corr/abs-1711-05144,DBLP:journals/corr/abs-1711-08513}, which conditions on any and all combination of covariates, but as we discussed, this requires a significant amount of data.  Ultimately, all of these approaches leave the question of how to interpret conditioning on different terms.

We hope that by forcing the definition of the fairness metric to specify the conditioned variables, practitioners consider the distribution of their evaluation data, how it was sampled, and the implications.
Unfortunately, this still does not diminish the importance of considering how the dataset is sampled or generated, and requires careful consideration by practitioners in determining this procedure.

\section{Correlation Loss}
\label{sec:mindiff}
Although previous work has laid out multiple techniques for training models with fairness metrics as part of the objective, these approaches generally come with notable engineering concerns.  Here, we present a new, lightweight approach for improving the desired fairness metrics more effectively than previous approaches.  

One perspective on group fairness metrics is that the output distribution should match across groups, possibly after reweighting or resampling the data \cite{EOadv,laftr}.  Notably, \cite{DBLP:journals/corr/abs-1801-07593} focuses on the prediction $\hat{y}$ rather than an intermediate layer. One way we could conceptualize this goal is to compare the distributions of $\hat{y}$ between groups and encourage low mutual information or KL divergence.  Here, we take a simplified view of that by minimizing the absolute correlation between $\hat{y}$ and the group membership $s$:
\begin{align}
\min_f \left[\sum_{(\mathbf{x}_i,y_i) \in \mathcal{X}} L(y_i, f(\mathbf{x}_i))\right] + \lambda |{\rm Corr}_\negset|
\end{align}
where
\begin{align*}
Corr_\negset = \frac{(\sum_{\mathbf{x}_i \in \negset} f(\mathbf{x}_i) - \mu_{\hat{y}})
                         (\sum_{s_i \in \negset} s_i - \mu_{s})}{\sigma_{\hat{y}} \sigma_s}
\\\mu_{\hat{y}} = \frac{1}{|\negset|}\sum_{\mathbf{x}_i \in \negset} f(\mathbf{x}_i)
\quad\quad \mu_s = \frac{1}{|\negset|}\sum_{s_i \in \negset} s_i
\\\sigma_{\hat{y}}^2 = \frac{1}{|\negset|}\sum_{\mathbf{x}_i \in \negset} (f(\mathbf{x}_i) - \mu_{\hat{y}})^2
\\ \sigma_s^2 = \frac{1}{|\negset|}\sum_{s_i \in \negset} (s_i - \mu_s)^2
\end{align*}
Following \cite{EOadv,laftr}, we use $\negset = \{x_i \in \mathcal{X} | y_i < \tau\}$ in order to optimize for equality of opportunity; this could be extended to other reweighting or resampling schemes to address other metrics as in \cite{laftr}.  In practice, we use $\tilde{\mathcal{X}}^- \subset \negset$, which is a mini-batch of examples sampled from $\negset$. This follows a similar pattern as previous adversarial approaches in adding a penalty based on the distribution of the output, but unlike all of the adversarial approaches, no training of an adversary is required, which we find to greatly improve stability in practice.  While minimizing this term does not provably minimize the fairness metric, we find we get good results in practice, as we will show below.

\section{Improvements in Practice}
\begin{figure*}
    \centering
    \begin{subfigure}[b]{0.24\textwidth}
        \includegraphics[width=\textwidth]{figs/groupone_linear.pdf}
        \caption{Linear}
        \label{fig:linear_groupone}
    \end{subfigure}
    \begin{subfigure}[b]{0.24\textwidth}
        \includegraphics[width=\textwidth]{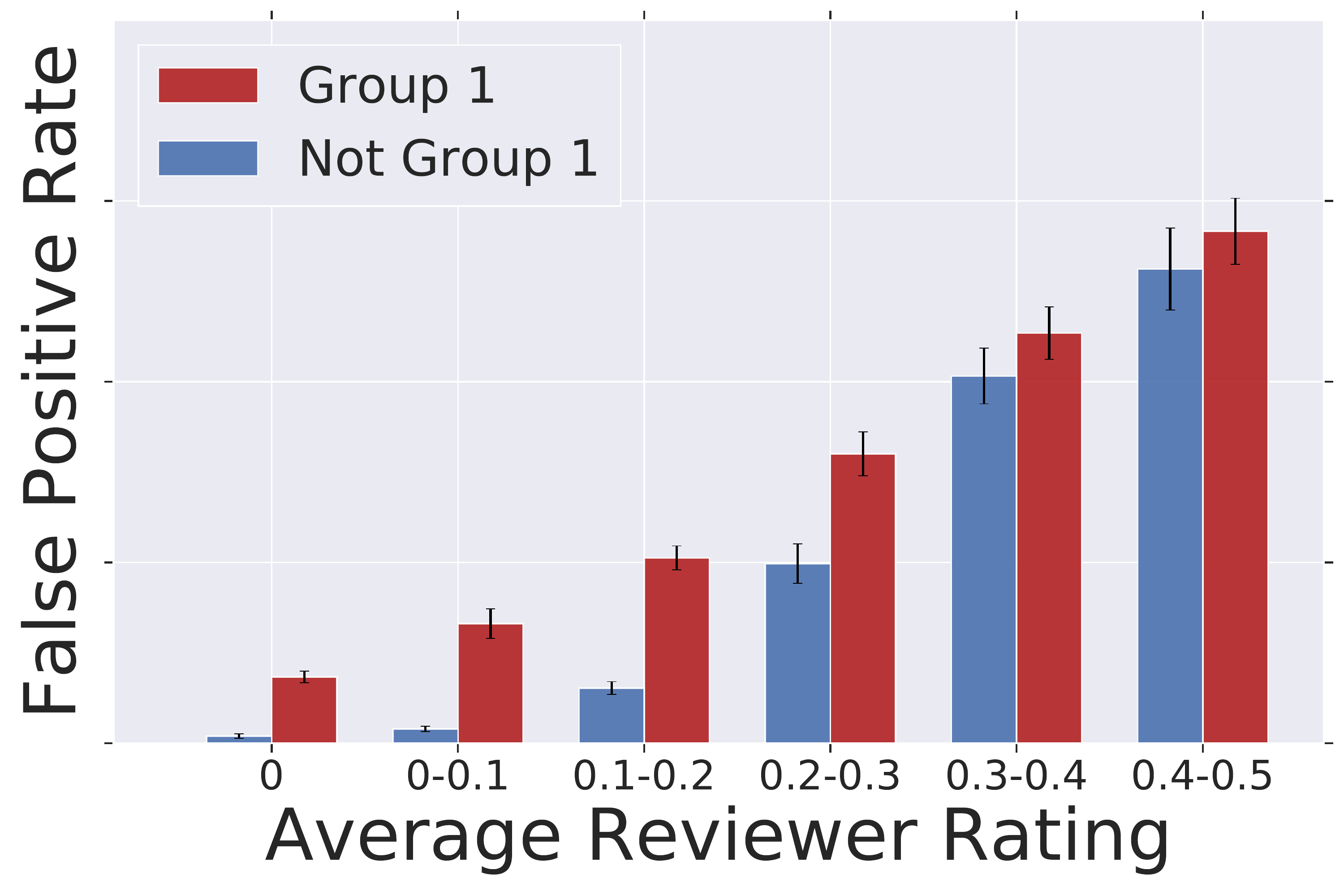}
        \caption{DNN}
        \label{fig:dnn_groupone}
    \end{subfigure}
        \begin{subfigure}[b]{0.24\textwidth}
        \includegraphics[width=\textwidth]{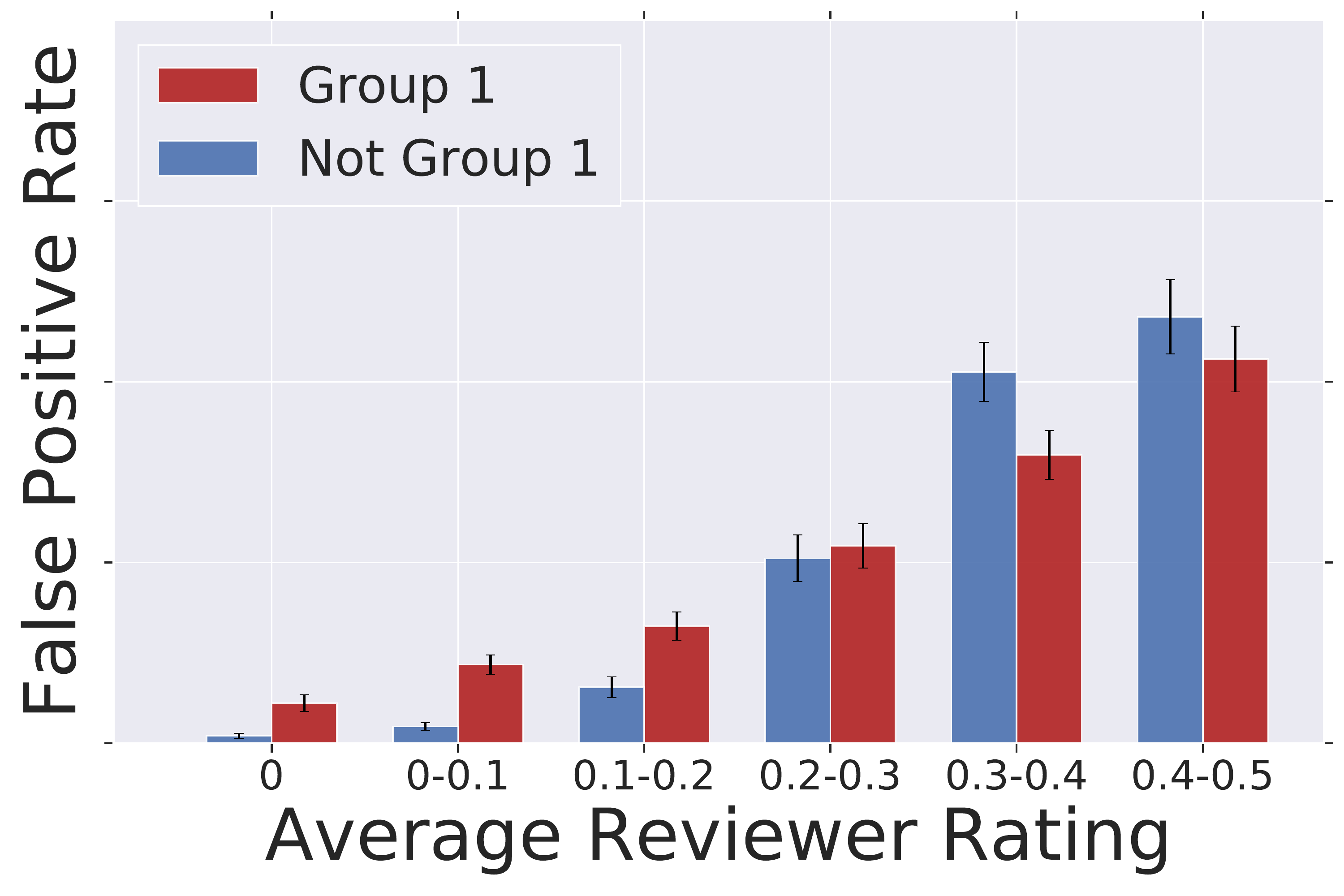}
        \caption{DNN with Adversary}
        \label{fig:dnn_adv_groupone}
    \end{subfigure}
    \begin{subfigure}[b]{0.24\textwidth}
        \includegraphics[width=\textwidth]{figs/groupone_mindiff_groupone.pdf}
        \caption{DNN with \mindifflossshort}
        \label{fig:dnn_mindiff_groupone}
    \end{subfigure}
    \caption{Improvements for \groupone \users.}
    \label{fig:groupone_improvement}
\end{figure*}

\begin{figure*}
    \centering
    \begin{subfigure}[b]{0.24\textwidth}
        \includegraphics[width=\textwidth]{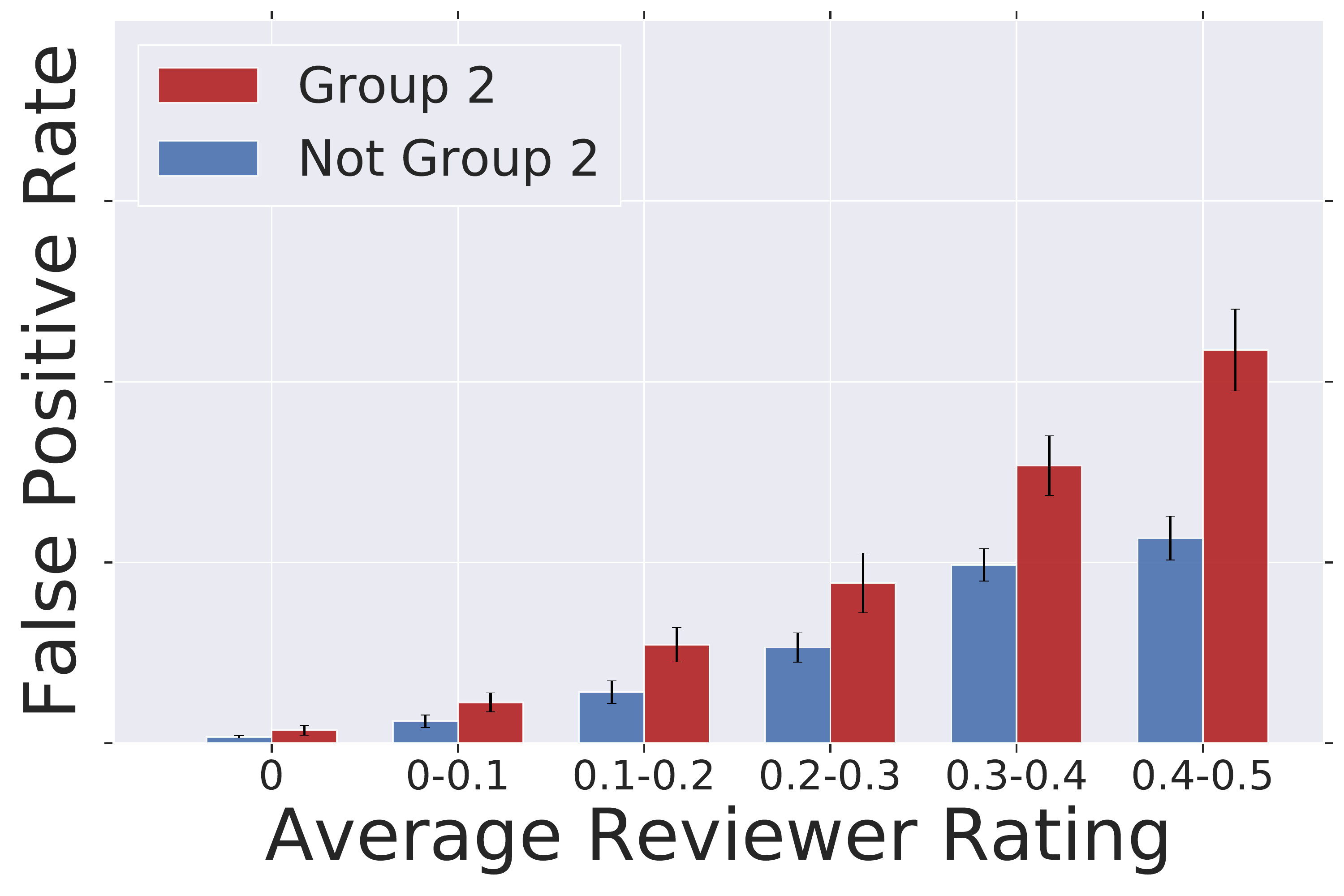}
        \caption{Linear}
        \label{fig:linear_grouptwo}
    \end{subfigure}
    \begin{subfigure}[b]{0.24\textwidth}
        \includegraphics[width=\textwidth]{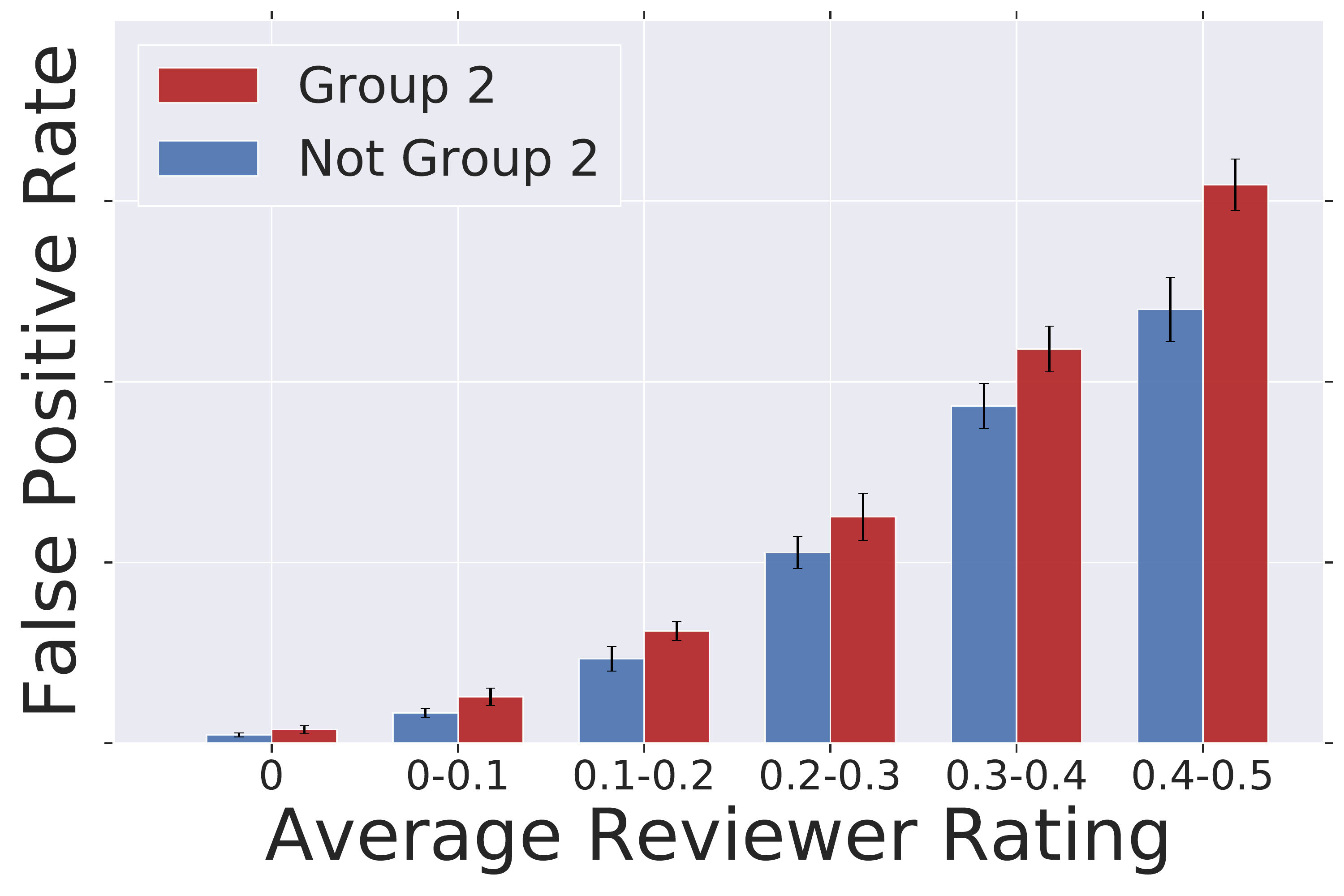}
        \caption{DNN}
        \label{fig:dnn_grouptwo}
    \end{subfigure}
    \begin{subfigure}[b]{0.24\textwidth}
        \includegraphics[width=\textwidth]{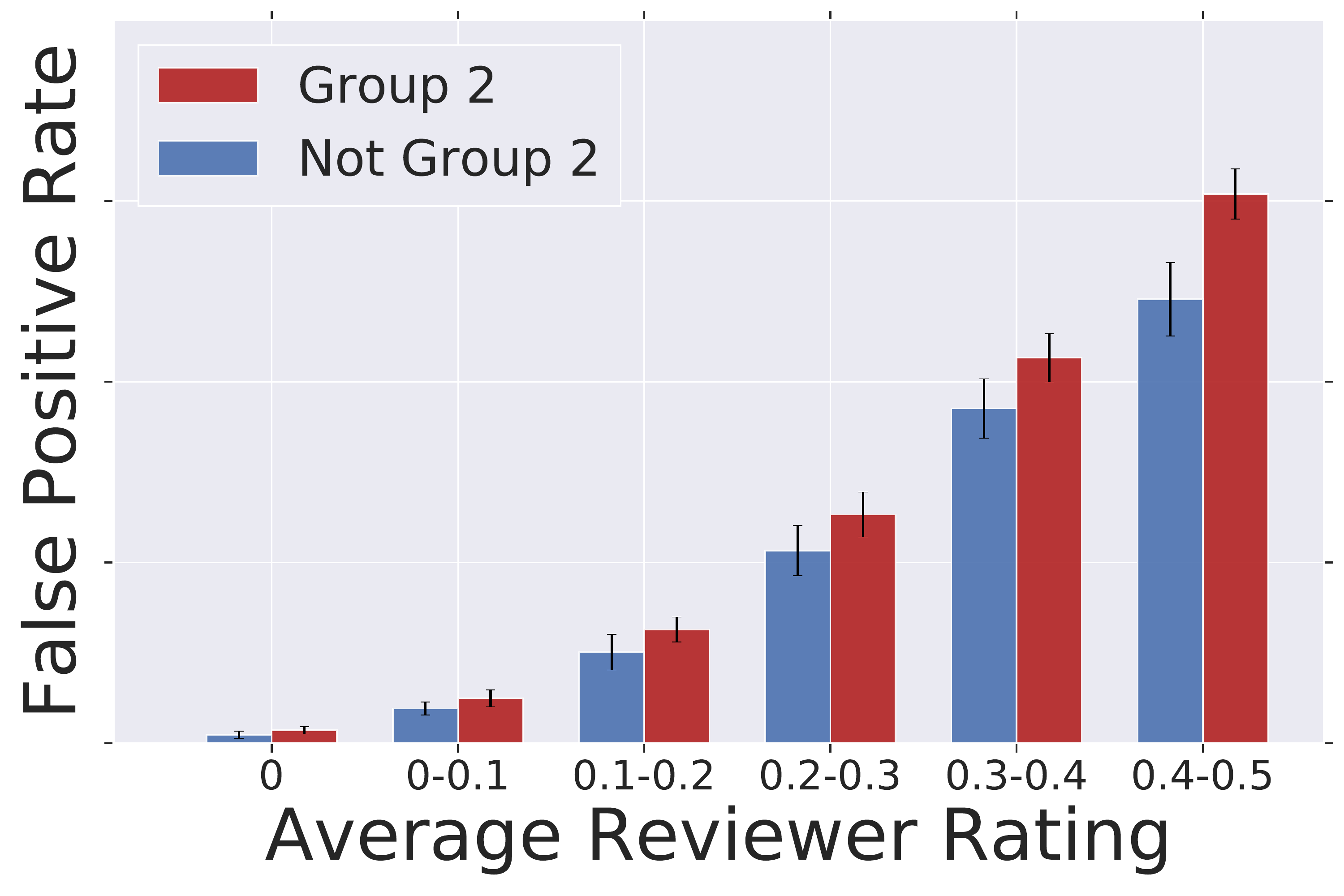}
        \caption{DNN + \groupone \mindifflossshort}
        \label{fig:dnn_mindiff_grouptwo}
    \end{subfigure}
    \begin{subfigure}[b]{0.24\textwidth}
        \includegraphics[width=\textwidth]{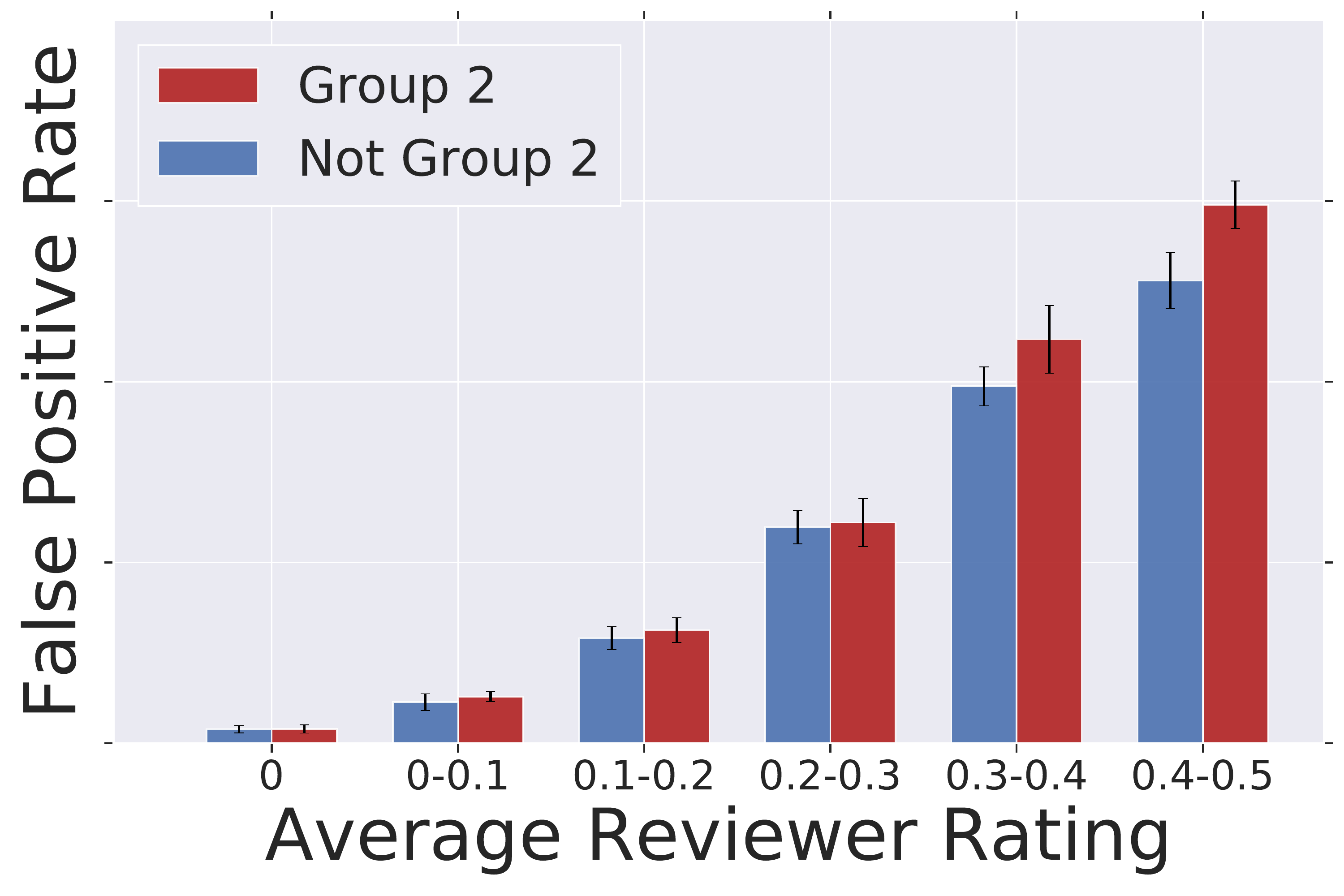}
        \caption{DNN + Both \mindifflossshort}
        \label{fig:dnn_bothmindiff_grouptwo}
    \end{subfigure}
    \caption{Improvements for \grouptwo \users.}
    \label{fig:grouptwo_improvement}
\end{figure*}

In practice, we seek to improve both the general equality of opportunity metric as well as the conditional equality metric.   All results here build on the description of the application setting, metrics and methods above.  In particular, we explore the incremental process by which we worked to improve this classifier, and how each change effects the end metrics.  Results are summarized in Figures \ref{fig:groupone_improvement} and \ref{fig:grouptwo_improvement} for \groupone and \grouptwo respectively, but we will walk through each of the results below.  Error bars are based on training the model 10 times and averaging results.  We find there are a number of trade-offs and directions for future work.

\paragraph{Model Capacity}
As discussed previously, the original model was a linear model, and we observe a significant gap in FPR between the sensitive subgroup and the rest of the data for both \groupone and \grouptwo, as seen in Figures \ref{fig:linear_groupone} and \ref{fig:linear_grouptwo}.  The first step we took was to change from a linear classifier to a DNN with a single hidden layer.  Previous work \cite{chen2018my} has suggested that theoretically model capacity can be a driver of disparities in accuracy.  As can be seen in Figure \ref{fig:dnn_groupone}, the move to DNN decreases the FPR for \groupone, as well as decreasing the \fprgap from \ratio{2.62} to \ratio{1.44}.  While this is good to see, this does not always hold true: we observe the FPR \emph{increases} for \grouptwo and \notgrouptwo users\footnote{Note, \notgroupone and \notgrouptwo are different in that they are based on different methods for getting demographic data $s$.}; the \fprgap decreases from \ratio{1.76} to \ratio{1.25}.  As such, while increasing model capacity may help in some cases, it does not necessarily improve accuracy everywhere.

\paragraph{Adversarial training}
Building on the DNN model, we next considered how well adversarial training can improve the \fprgap.  We follow an approach similar to \citet{EOadv} of training an additional head taking as an input the last hidden layer of the model and trying to predict the sensitive attribute $s$ while the model tries to learn a representation that is independent of $s$; we use only data for which $y < \tau$, as we are concerned with the FPR.  As we see in Figure \ref{fig:dnn_adv_groupone}, this significantly decreases the \fprgap from \ratio{1.44} to \ratio{1.04} and simultaneously decreases the FPR for \groupone, despite that not being an explicit objective.

\paragraph{Correlation Loss}
As mentioned previously, adversarial training has been well studied and we see has strong performance, but from an engineering perspective is challenging due to its instability during training.  As a result, we pursued  \mindiffloss to stabilize training.  We observe in Figure \ref{fig:dnn_mindiff_groupone} that using \mindiffloss keeps the \fprgap approximately the same (\ratio{1.05}).  
The practical benefits of keeping a low  \fprgap while improving stability is highly valuable in practice.

\paragraph{Transfer across Groups}
One idea that has been debated in the literature \cite{laftr} is if and when there is transfer learning of fairness across groups.  We consider here how the application of \mindiffloss to \groupone effects the \fprgap for \grouptwo \users.  In Figure \ref{fig:dnn_mindiff_grouptwo} we observe a very slight improvement in the \fprgap, bringing it down from \ratio{1.37} to \ratio{1.31}.  

\paragraph{Improving for Multiple Groups}
Finally, we look at if we can simultaneously improve the \fprgap for both \groupone and \grouptwo \users simultaneously.  To do this, we add two different \mindiffloss terms to the DNN training, one for each group.  As we see in Figure \ref{fig:dnn_bothmindiff_grouptwo}, we are able to improve the \fprgap for \grouptwo to \ratio{1.11}.  Unfortunately, we do not find that this decreases the FPR for \grouptwo; we believe focusing on making the model more inclusive by improving the accuracy not just decreasing the gap is an important step for future work \cite{beutel2017beyond,chen2018my}.

\section{Future Directions}
\label{sec:limits}
We have focused on how to improve an individual model that directly effects the user experience, but by no means does every use of machine learning fit into these settings. To expand the applicability, we believe there are a number of areas that deserve more research attention.

\myparagraph{Human raters:} This work, like most of the \mlfairness literature, assumes that the labels are unbiased.  We believe more attention is needed to understand if and when there is bias in crowd-sourced ratings, and how to remove it.

\myparagraph{Binary actions:} We only consider the case where we are taking binary actions directly against the examples (at the known threshold).  When the prediction is treated as a continuous score or used in conjunction with other signals, it becomes harder to evaluate the effect on the user experience \cite{composefairness} and further research is needed in this direction.

\myparagraph{Observed Examples:} We evaluate our system against the examples currently in the system.  However, that distribution is of course affected by the system's previous performance.  Use cases that were previously not well supported may be underrepresented in our sample.  Unfortunately, we do not know of a way to infer the distribution of examples that would exist under a different previous system.  As a result, for the time being, we focus on evaluating and improving metrics for the current state, with the belief that it will improve the performance of the system for the sensitive subgroups and we can continue to evaluate the performance as the subgroup evolves.
Further research, along the lines of \cite{delayedimpact}, would be valuable to better understand the long term evolution of the system.

\section{Discussion}
In this work we have offered details on how applying \mlfairness principles to a production classifier fares.  In particular, we have explored how equality of opportunity depends on how the data is sampled, and how different groups can have notably different distributions of data.  To address issues with these distributional differences, we offered a general evaluation approach that takes into account example difficulty.  
Further, we have offered a new mechanism, \mindiffloss, for improving \mlfairness metrics that we find to be more stable than adversarial training.  We demonstrate the ability of these approaches to improve the \fprgap for two different groups in a production classifier and offer analysis of how the model performance is effected by these different training procedures.

\noindent {\footnotesize \textbf{Acknowledgements:} We would like to thank Margaret Mitchell for her valuable feedback on this paper. }

\newpage
\newpage
\clearpage
{
\footnotesize
\bibliographystyle{aaai}
\bibliography{alex}
}

\end{document}